\documentclass[letterpaper, 10 pt, conference]{ieeeconf}  
\IEEEoverridecommandlockouts                              

\usepackage{times}

\usepackage[hidelinks]{hyperref}
\usepackage[utf8]{inputenc}
\usepackage[small]{caption}
\usepackage{graphicx}

\usepackage{language}
\usepackage[spanish,english]{babel}
\usepackage{color}
\usepackage{xspace}
\usepackage{url}
\usepackage{amsmath}
\newtheorem{definition}{Definition}
\usepackage[noend]{algpseudocode}
\usepackage{algorithm}

\usepackage{tabularx,booktabs}
\newcolumntype{Y}{>{\centering\arraybackslash}X}
\newcolumntype{Z}{>{\raggedleft\arraybackslash}X}
\newcolumntype{T}{>{\raggedright\arraybackslash}X}
\newcolumntype{s}{>{\hsize=.75\hsize}Y}

\newcommand{\lamafirst}{{\sc lama-first}\xspace}

\newcommand{\replan}{{\sc replan}\xspace}
\newcommand{\clo}{{\sc clo}\xspace}

\newcommand{\rooms}{{\sc Rooms}\xspace}
\newcommand{\dialog}{{\sc Dialog}\xspace}
\newcommand{\cooking}{{\sc Cooking}\xspace}

\newcommand{\placefigure}[2]{%
  \vfill
  \begin{center}
    \includegraphics[scale=#1]{#2}
  \end{center}
  \vfill
}

\newcommand{\example}[1]{{\bf Example.} \emph{#1}}
\newcommand{\cexample}[1]{\emph{#1}}

\title{\LARGE \bf Intelligent Execution through Plan Analysis}

\author{Daniel Borrajo$^{1}$\thanks{$^{1}$Consultant. On leave from Universidad Carlos III de Madrid} and
  Manuela Veloso$^{2}$\thanks{$^{2}$On leave from Carnegie Mellon University, School of Computer Science}\\
  J.P.Morgan AI Research, New York, NY (USA)\\
  \{daniel.borrajo,manuela.veloso\}@jpmchase.com}

\begin{document}

\maketitle

\thispagestyle{empty}
\pagestyle{empty}

\begin{abstract}
  Intelligent robots need to generate and execute plans. In order to deal with the complexity of real environments, planning makes some assumptions about the world. When executing plans, the assumptions are usually not met. Most works have focused on the negative impact of this fact and the use of replanning after execution failures. Instead, we focus on the positive impact, or opportunities to find better plans. When planning, the proposed technique finds and stores those opportunities. Later, during execution, the monitoring system can use them to focus perception and repair the plan, instead of replanning from scratch. Experiments in several paradigmatic robotic tasks show how the approach outperforms standard replanning strategies.
\end{abstract}

\section{Introduction}

One of the key intelligent capabilities of robots is their ability to generate plans.  Automated planning systems
generate plans as sequences of actions from an initial state to achieve a set of goals~\cite{planning-book}. In order to
generate those plans, planners assume the models of their environment they take as input are the right ones. However,
robotic environments are dynamic and change throughout the execution of plans in ways unexpected by the planning
models~\cite{VelosoBCR15}. For instance, actions preconditions and effects might not be the right ones or they might
evolve; new facts might appear that were not defined in the domain model; objects can appear or disappear; facts can
change their truth value; or goals can become irrelevant or new goals might appear~\cite{rogue-plan-exec}.

Thus, executives continuously monitor the execution for plan validity. If failures are detected (the effects were not
the expected ones, and/or they can avoid achieving the goals with the current plan), then they replan. Most executives
only check for failures. However, there is a second type of unexpected result from the actions execution, facts that
represent opportunities to improve the plan in execution.

In general, in order to improve the execution, there are three levels of information that can be exchanged between a
planner and the execution system:

\begin{enumerate}
\item plan: this is the simplest and most used approach. The planner provides no annotation to the executive, so when
  the execution does not proceed as expected, it must replan. Replanning can be done from
  scratch~\cite{FF-REPLAN-ICAPS07} or by taking into account the previous
  plan~\cite{errt,lpgfoxetal06,workshop-icaps12-errtplan}.
\item plan plus subgoaling structure: the planner attaches to each action the goals and subgoals that it can achieve
  directly or indirectly. It also includes precedence constraints among actions. This kind of plan annotation increases
  the ability of the executive to reason about changes to the expected behavior of the plan. For instance, it can detect
  whether the changes affect the future execution of the plan, and replan if so~\cite{Fritz07,VelosoPC98}.
\item plan plus rationale on why the planner generated this plan: the plan is also annotated with other relevant
  planning-related information that can yield to better decisions in terms of how to repair a plan~\cite{mljournal}.
\end{enumerate}

In recent work, we used the plan rationale to reason about {\it static} opportunities~\cite{icaps21}; that is, facts
that cannot change by the robot's actions, such that if their value changes in the environment, it can yield to better
plans by replanning. Instead, in this paper, we use the subgoaling structure to reason about {\it dynamic}
opportunities, facts that can be changed by the robot actions. The paper provides as contributions novel techniques to
solve the following challenges: (1) how can the system compute dynamic opportunities at planning time; (2) how can they
be detected when they appear during execution; and (3) how can the executive use them efficiently. First, opportunities
are computed by using the causal links among actions of the plan. Causal links capture the information on what actions
achieve some facts that are needed as preconditions of other actions~\cite{UCPOP}. Second, opportunities serve two
purposes to the executive: detect when an opportunity arises; and focus the perception system only on those changes in
the state that matter with respect to finding those opportunities. This second aspect is crucial in robotic applications
where the perception system can focus on many different aspects of the environment. Third, opportunities allow the
executive to detect which components of the plan are not needed anymore given the appearance of those opportunities, so
that it can repair the plan without performing expensive replanning.

\section{Related Work}

Some previous works have defined different methods for performing plan repair, for case-based
planning~\cite{mljournal,acmsurvey15} or for making changes to an existing
plan~\cite{errt,lpgfoxetal06,workshop-icaps12-errtplan}.  In these works, there is no notion of opportunities; they just
focus on the repair task. Other works focused on monitoring for optimal plans~\cite{Fritz07}. The planner provides part
of the rationale built for generating the plan and it serves the purpose of maintaining optimality on the execution of a
plan. This work requires to work in optimal planning, which is unrealistic in many robotic applications.

Opportunities have also been studied in cognitive science for human
reasoning~\cite{seifert2001opportunism}.
These works focused on which pending goals to pursue when
opportunities arise, instead of using opportunities for guiding perception, and performing repair of a plan. Also, these
works were not implemented.

In relation to robotics, there has been some work related to opportunism in navigation tasks, but opportunism was
defined as solving the Team Orienteering Problem~\cite{Thakur}, as alternative actions that could be more desirable for
the robot~\cite{kruse2010}, or it was used to compute alternative routes for navigation~\cite{Cefalo2020,chaves2014}.
In general, robotics-related use of opportunities has focused in a single task, defining task-dependent models. In our
case, we present a domain-independent technique.  Other domain-independent planning related works have focused on
pre-defined goals that could be achieved in case of underused resources or resources
uncertainty~\cite{Cashmore17,coles12};
or new goals to be pursued (with little or no re-planning
capabilities)~\cite{Lawton04}. In those cases, opportunities are pending goals that are not part of the causal structure
of the current plan, while our opportunities are part of the causal structure.

\section{Background}

A classical planning task is a tuple $\Pi=\{F,A,I,G\}$, where $F$ is a set of facts, $A$ is a set of instantiated
actions, $I\subseteq F$ is an initial state represented by the set of facts that are true initially, and $G\subseteq F$
is a set of goals, facts that should be true at the end.  Each action $a\in A$ is described by a set of preconditions
(pre($a$))
and a set of effects (eff($a$)).
The definition of each action might also include a cost $c(a)$ (the
default cost is one).
The planning task should generate as output a plan, sequence of actions $\pi=(a_1,\ldots,a_n)$ such that if applied in
order from the initial state $I$ would result in a state $s_n$, where goals are true, $G\subseteq s_n$.  Plan cost is
commonly defined as: $C(\pi)=\sum_{a_i\in\pi} c(a_i)$.

We will use the concept of causal link from partial-order planning~\cite{UCPOP} to define opportunities.

\begin{definition}[Causal link]
  A causal link is a tuple $\langle a_p, f, a_c\rangle$ that specifies that action $a_p$ (producer) achieves an effect
  $f\in F$ that is needed as a precondition of action $a_c$ (consumer).
\end{definition}

$\cal L$ will be the set of causal links from a plan. Since there might be more than one instance of the same action in
the plan, we will use a vector for $\cal L$, where each position represents a step in the plan, and will hold a set of
causal links on that step.

\begin{figure}[hbt]
\placefigure{0.3}{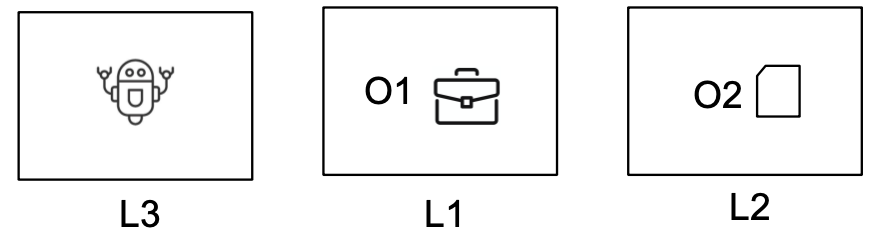}
\caption{Example of the \rooms domain with one robot, two objects and three rooms. The goal is to grab all objects.}
\label{fig:office}
\end{figure}

\example{Suppose we have a robot that is taking actions in an environment with rooms and objects as shown in
  Figure~\ref{fig:office}. We named the domain \rooms. Let us assume it can execute three actions: {\tt move} from one
  room to another, {\tt prepare} an object for grasping and {\tt grasp} the object. In the example, it operates in an
  environment composed of three rooms - locations - (L1, L2 and L3), where there are two objects (O1 at L1, O2 at
  L2). Initially, it is at L3 and the goal is to hold both objects.}

\cexample{Table~\ref{tab:actions} shows schematically the lifted actions in this domain. Names preceded by a question
  mark are variables. The actual actions in $A$ are computed by instantiating the variables with the objects in the
  initial state. As mentioned before, actions are modeled with a set of preconditions that need to be true before
  applying them, and a set of effects that are expected to be true (add effects) or false (del effects, preceded by
  $\neg$) after executing the action. In this domain, {\tt at-robot(?l)} means that the robot is at a given room {\tt
    ?l}, {\tt prepared(?o)} means that an object is ready to be grasped, and {\tt holding(?o)} means that object {\tt
    ?o} is held by the robot.}

\begin{table}[hbt]
  \caption{Actions in the \rooms domain.}
  \label{tab:actions}
  \begin{tabularx}{0.47\textwidth}{*{3}{T}}
    \toprule
    Action & Preconditions & Effects\\ \midrule
    move(?l1,?l2) & at-robot(?l1) & at-robot(?l2)\\
    & & $\neg$at-robot(?l1)\\ \midrule
    prepare(?o,?l) & at-object(?o,?l) & prepared(?o)\\
    & at-robot(?l) & \\ \midrule
    grasp(?o,?l) & at-object(?o,?l) & holding(?o)\\
    & at-robot(?l) & $\neg$at-object(?o,?l)\\
    & prepared(?o) &\\
    \bottomrule
  \end{tabularx}
\end{table}

\cexample{A potential causal link in this domain would be:}

\cexample{$\langle$move(L3, L1), at-robot(L1), prepare(O1, L1)$\rangle$}

\cexample{The action {\tt move(L3, L1)} achieves the proposition {\tt at-robot(L1)} that is needed by action {\tt
    prepare(O1, L1)} as one of its preconditions.}

\section{Computation of Causal Links Opportunities}

Given a planning task and a plan, the computation of Causal Links based Opportunities, \clo, follows
Algorithm~\ref{algo}. It computes the set of causal links $\cal L$ from which it extracts the opportunities and that
will also serve to perform intelligent repair. It computed the set from a valid plan and it starts from the end of the
plan in reverse order, to there cannot be threats in $\cal L$. Threats are actions that could potentially be executed
between the producer and consumer actions are executed and that would delete the fact of the causal link. The algorithm
starts by initializing $P$, a set of pairs where each pair is composed of a pending goal (the algorithm did not find yet
an action that can achieve it), and the action that needs that goal to be true (consumer). At start, pairs will be
formed by the goals of the problem and a virtual action $A_\infty$. Then, for each action in the plan (starting from the
end), and each effect of that action, if that effect is needed by any pending goal, it creates a causal link and removes
the corresponding pending goal from $P$. The final step in the loop inserts the preconditions of the action into
$P$. The algorithm returns the set of causal links that will be used in monitoring. The set of opportunities $\cal O$ is
the set of all propositions involved in at least one causal link.

\begin{algorithm}[hbt]
\caption{{\sc Causal Links Opportunities}, \clo$(\Pi, \pi)$}
\label{algo}
\begin{algorithmic}[1]
  \Require planning task $\Pi=\{F,A,I,G\}$, plan $\pi$
  \Ensure set of causal links $\cal L$, opportunities, $\cal O$
  \State {\it Init:} $\forall i\in[1..|\pi|-1], {\cal L}(i)\gets\emptyset$
  \State {\it Init:} $P\gets\{\langle g, A_\infty\rangle: \forall g\in G\}$; $i\gets 1$
  \ForAll{$a\in$reverse($\pi$)}
  \ForAll{$f\in\text{eff}(a)$}
  \ForAll{$p=\langle f, a_c\rangle\in P$}
  \State ${\cal L}(i)\gets{\cal L}(i)\cup\{\langle a, f, a_c\rangle\}$
  \State $P\gets P\setminus \langle f, a_c\rangle$
  \EndFor
  \EndFor
  \State $P\gets P\cup\{\langle p, a\rangle: p\in\text{pre}(a)\}$
  \State $i\gets i + 1$
  \EndFor
  \State ${\cal O}\gets\{f \mid \langle a_p, f, a_c\rangle\in\cal L\}$
  \State \Return $\cal L, O$
\end{algorithmic}
\end{algorithm}

\example{The causal links associated to the plan to solve the problem in Figure~\ref{fig:office} are represented in
  Table~\ref{tab:cl}, where the first column shows the plan actions.}

\begin{table}[hbt]
  \caption{Causal links of the plan to solve problem in Figure~\ref{fig:office}.} 
  \label{tab:cl}
  \begin{tabularx}{0.47\textwidth}{>{\hsize=0.2\hsize}X>{\hsize=0.5\hsize}X>{\hsize=0.5\hsize}X>{\hsize=0.5\hsize}X} 
    Step & Producer & Proposition & Consumer\\ \midrule
    1 & move(L3, L1) & at-robot(L1) & prepare(O1, L1)\\
    & move(L3, L1) & at-robot(L1) & grasp(O1, L1)\\
    & move(L3, L1) & at-robot(L1) & move(L1, L2)\\  \midrule
    2 & prepare(O1, L1) & prepared(O1) & grasp(O1, L1)\\ \midrule
    3 & grasp(O1, L1) & holding(O1) & $A_\infty$\\ \midrule
    4 & move(L1, L2) & at-robot(L2) & prepare(O2, L2)\\
    & move(L1, L2) & at-robot(L2) & grasp(O2, L2)\\ \midrule
    5 & prepare(O2, L2) & prepared(O2) & grasp(O2, L2)\\ \midrule
    6 & grasp(O2, L2) & holding(O2) & $A_\infty$\\ \bottomrule
  \end{tabularx}
\end{table}

\cexample{The opportunities in this example would be the set:}

\cexample{$\cal O =$\{{\tt at-robot(L1), prepared(O1), holding(O1), at-robot(L2), prepared(O2), holding(O2)}\}}

\section{Intelligent Monitoring}

The goal of the monitoring and execution system is to execute actions, check if a dynamic opportunity has appeared and
modifying the plan if needed. Algorithm~\ref{execution} performs the execution-monitoring loop. It first computes the
plan ($\pi$), and the causal-links ($\cal L$) and opportunities $\cal O$. Then, it loops while it did not achieve the
goals and there are still actions to be executed. It executes the first action of the remaining plan, and senses the
environment for the new state. The sensing action takes as input the opportunities and the action, since it will focus
its attention on those facts plus the effects of the action. For each opportunity in the new state, it calls the
\Call{repair}{} algorithm.
Then, if replanning is needed due to a failed execution (preconditions of next actions are not met), then a new plan and
opportunities are computed by calling the planner and the \clo algorithm. The algorithm returns True if the goals were
achieved and False otherwise.

\begin{algorithm}[hbt]
\caption{{\sc Opportunities Plan Execution}($\Pi$)}
\label{execution}
\begin{algorithmic}[1]
  \Require planning task $\Pi=\{F,A,I,G\}$
  \Ensure boolean (solved or not)
  \State {\it Init:} $\pi\gets$\Call{Plan}{$\Pi$}; ${\cal L, O}\gets$\Call{\clo}{{$\Pi,\pi$}}; $s\gets I$
  \While{$\pi\not=\emptyset$ AND $G\not\subseteq s$}
  \State $a\gets$\Call{pop}{$\pi$}
  \State \Call{Execute}{$a,\Pi$}
  \State $s'\gets$\Call{Sense}{$a,\cal O$}
  \ForAll{$o\in s'\cap{\cal O}$}
  \State $\pi, {\cal L, O}\gets$\Call{Repair}{$o,\pi,{\cal L, O}$}
  \EndFor
  \If{\Call{NeedReplanning}{$s',\pi$}}
  \State $\pi', {\cal L, O}\gets$\Call{PlanOpportunities}{$\{F,A,s',G\},\pi,\cal L$}
  \If{$\pi'\not=\emptyset$}
  \State $\pi\gets\pi'$
  \Else \State \Return False 
  \EndIf
  \Else \State ${\cal L}\gets\text{rest}({\cal L})$
  \EndIf
  \State $s\gets s'$
  \EndWhile
  \If{$G\subseteq s$}
  \State \Return True
  \Else \State \Return False
  \EndIf
\end{algorithmic}
\end{algorithm}

Algorithm~\ref{repair} presents the pseudo-code of the repair strategy. In a nutshell, it removes the opportunity $o$
from all the causal links where it was needed. If the set of causal links of any action becomes empty, the action is no
longer needed.  Algorithm~\ref{removeactions} removes all such actions iteratively. The same action can appear multiple
times in a plan, so the algorithm only removes the first one found in the remaining plan and removes actions backwards
from that action.

\begin{algorithm}[hbt]
\caption{{\sc Repair}($o,\pi,{\cal L, O}$)}
\label{repair}
\begin{algorithmic}[1]
  \Require found opportunity, $o$, current plan $\pi$, causal links $\cal L$
  \Ensure new plan $\pi$, new causal links $\cal L$, opportunities, $\cal O$
  \Repeat
  \State {\it Init:} changed?$\gets$False; $i\gets 0$
  \Repeat
  \ForAll{$l=\langle a_p, o, a_c\rangle\in{\cal L}(i)$}
  \State ${\cal L}(i)\gets {\cal L}(i)\setminus\{l\}$
  \State $a\gets a_p$
  \State changed?$\gets$True
  \EndFor
  \If{${\cal L}(i)=\emptyset$}
  \State $\pi, {\cal L}\gets$\Call{RemoveActions}{$i, \pi, {\cal L}$}
  \EndIf
  \State $i\gets i + 1$
  \Until{changed? or $i=|{\cal L}|$}
  \Until{not(changed?)}
  \State ${\cal L}\gets\{l: l\in{\cal L}, l\not=\emptyset\}$
  \State ${\cal O}\gets\{f \mid \langle a_p, f, a_c\rangle\in\cal L\}$
  \State $\pi'\gets\{a: a\in\pi, a\not=\emptyset\}$
  \State \Return $\pi', \cal L, O$
\end{algorithmic}
\end{algorithm}

\begin{algorithm}[hbt]
\caption{{\sc RemoveActions}($i, \pi,{\cal L}$)}
\label{removeactions}
\begin{algorithmic}[1]
  \Require position of $a$ in $\pi$ $i$, current plan $\pi$, causal links $\cal L$
  \Ensure new plan $\pi$, new causal links $\cal L$
  \State {\it Init:} $j\gets|\pi|$; $A^-\gets\{i\}$
  \Repeat
  \State found?$\gets$False
  \State $i\gets\text{pop}(A^-)$
  \State $a\gets \pi(i)$
  \State $k\gets i$ 
  \Repeat
  \ForAll{$l=\langle a_p, f, a_c\rangle\in {\cal L}(k)$}
  \If{$a_c=a$}
  \State ${\cal L}(k)\gets {\cal L}(k)\setminus\{l\}$
  \EndIf 
  \If{$a_p=a$}
  \State found?$\gets$True
  \EndIf
  \EndFor
  \If{${\cal L}(k)=\emptyset$}
  \State $A^-\gets A^-\cup\{k\}$
  \State $\pi(k)\gets\emptyset$
  \EndIf
  \State $k\gets k-1$
  \Until{found? or $k=0$} 
  \Until{$A^-=\emptyset$}
  \State \Return $\pi, {\cal L}$
\end{algorithmic}
\end{algorithm}

\example{Suppose the robot starts executing the plan by applying the action {\tt move(L3, L1)}. If the new state only
  changes in the robot position, it continues with the execution since this was the expected state. If new facts become
  true but they are not opportunities, it also continues the execution. It will only react to state changes that could
  affect the causal links of the plan. If the new state includes changes in any of the opportunities, then it will try
  to repair the plan. Suppose that {\tt holding(O2)} becomes true after executing the action {\tt move(L3, L1)} because
  another agent has prepared and given the object to the robot. The {\sc repair} algorithm would find the first step
  where there is a causal link that uses that fact, step 6. The action {\tt grasp(O2, L2)} will achieve that fact, which
  is needed by the goals ($A_\infty$ action). Then, the algorithm removes the causal link in step 6 from $\cal L$. If
  the set of causal links at that step becomes empty, it means that the corresponding action, {\tt grasp(O2, L2)}, is
  not needed; that is, the action does not provide anything useful to the plan. This is what happens to step 6. Since
  the action is no longer needed, it analyzes backwards from that step which causal links produced some fact to be
  consumed by that action, because those causal links will also not be needed. We see that step 5 had only one causal
  link whose consumer action is {\tt grasp(O2, L2)}. Therefore, we can also remove that causal link, and proceed
  backwards. Since step 5 becomes empty, the algorithm also adds action {\tt prepare(O2, L2)} to the list of actions to
  be removed. Step 4 also included {\tt grasp(O2, L2)} as a consumer of a causal link, so that causal link is
  removed. However, the producer action, {\tt move(L1, L2)} still has other causal links, so it is not added to the list
  of actions to be removed yet. At this step, the remaining causal links are depicted in Table~\ref{tab:cl1}.}

\begin{table}[hbt]
  \caption{Causal links after removing two causal links.}
  \label{tab:cl1}
  \begin{tabularx}{0.47\textwidth}{>{\hsize=0.2\hsize}X>{\hsize=0.5\hsize}X>{\hsize=0.5\hsize}X>{\hsize=0.5\hsize}X}
    \toprule
    Step & Producer & Proposition & Consumer\\ \midrule
    2 & prepare(O1, L1) & prepared(O1) & grasp(O1, L1)\\ \midrule
    3 & grasp(O1, L1) & holding(O1) & $A_\infty$\\ \midrule
    4 & move(L1, L2) & at-robot(L2) & prepare(O2, L2)\\ \bottomrule
  \end{tabularx}
\end{table}

\cexample{The set of actions to be removed is now composed of {\tt grasp(O2, L2)} and {\tt prepare(O2, L2)}. The
  algorithm proceeds until there are no more actions to be removed.
  The algorithm returns: the new set of causal links, removing the corresponding steps; the new plan, removing all
  actions whose steps became empty after removing the causal links; and the new set of opportunities (propositions in
  the remaining causal links). The new set of causal links is shown in Table~\ref{tab:cl2} and the new plan would be:}
  
\cexample{$\pi=${\tt prepare(O1, L1)}, {\tt grasp(O1, L1)}.}
  
\cexample{The algorithm has obtained a new plan much shorter than the original one by removing actions that are no
  longer needed after finding an opportunity.}

\begin{table}[hbt]
  \caption{Causal links after removing the causal links that are no longer needed.}
  \label{tab:cl2}
  \begin{tabularx}{0.47\textwidth}{>{\hsize=0.2\hsize}X>{\hsize=0.5\hsize}X>{\hsize=0.5\hsize}X>{\hsize=0.5\hsize}X}
    \toprule
    Step & Producer & Proposition & Consumer\\ \midrule
    1 & prepare(O1, L1) & prepared(O1) & grasp(O1, L1)\\ \midrule
    2 & grasp(O1, L1) & holding(O1) & $A_\infty$\\ \bottomrule
  \end{tabularx}
\end{table}

\section{Experiments and Results}

In this section, we compare the performance of our proposed system, \clo, to that of \replan, a standard strategy that
replans every time there is some change between the perceived state and the expected state.  We have created versions of
three known robotic tasks: the \rooms domain defined before that covers navigation and grasping tasks; the \dialog
domain that covers human-robot interaction tasks; and the \cooking domain that covers home-related tasks. We have just
defined the barebones of those tasks to highlight that, even with minimalistic models, the technique introduced in this
paper outperforms standard replanning strategies. The more actions the plans have, the greater the benefits will be of
using our approach. For each domain, we generated four problems of increasing difficulty, named p-05, p-10, p-20 and
p-40 where we increase the number of objects and goals according to those numbers. The appearance of an opportunity at a
given time step depends on a probability that we have varied using the values 0.1, 0.2 and 0.5.

We used \lamafirst as the base planner
implemented in Fast Downward~\cite{fast-downward-jair}. Experiments were performed on a MacBook Pro, with processor 2.4
GHz Intel Core i5, 4Gb of RAM.  The time bounds were: 1800s for the complete planning-execution cycle; 500s for each
planning episode; and 300s for computing opportunities. These bounds were reached only by \replan in two cases.

\begin{figure*}[hbt]
  \hspace*{-0.3cm}
    \begin{tabular}{c@{\hspace*{-0.2cm}}c@{\hspace*{-0.2cm}}c}
      \includegraphics[scale = 0.415]{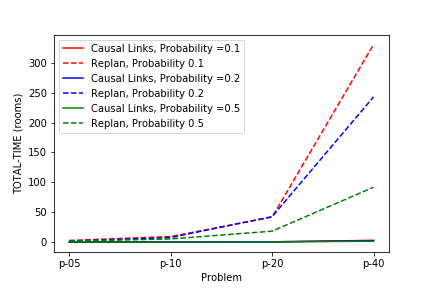} &
      \includegraphics[scale = 0.415]{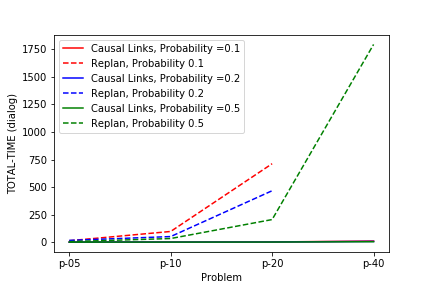} &
      \includegraphics[scale = 0.415]{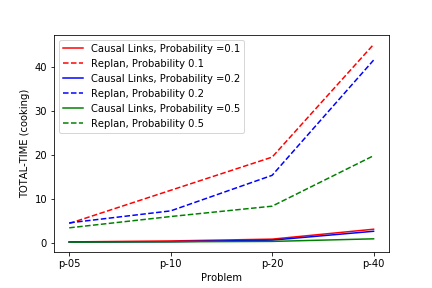}\\
      (a) & (b) & (c)
    \end{tabular}
    \caption{Total time in the (a) \rooms, (b) \dialog and (c) \cooking domains taken by \clo (replanning with
      opportunities) and \replan.}
\label{fig:exp}
\end{figure*}

\begin{table*}[hbt]
  \caption{Number of executed actions in the three domains.}
  \label{tab:num-actions}
\footnotesize
\begin{center}
\begin{tabular}{crrrrrrr}
  \toprule
  & & \multicolumn{2}{c}{\rooms} & \multicolumn{2}{c}{\dialog} & \multicolumn{2}{c}{\cooking}\\
  \cmidrule(lr){3-4} \cmidrule(lr){5-6} \cmidrule(lr){7-8}
Problem &  Probability &  \#actions \clo &  \#actions \replan &  \#actions \clo &  \#actions \replan &  \#actions \clo &  \#actions \replan \\
\midrule
   p-05 & 0.1 & 12.0 & 12.0 &    23.0 & 23.0 & 28.0 & 28.0 \\
   p-05 & 0.2 & 9.0 & 9.0 &    26.0 & 26.0   & 30.0 & 29.0 \\
  p-05 & 0.5 & 8.0 & 8.0 &    14.0 & 14.0    & 23.0 & 22.0 \\
  \midrule
   p-10 & 0.1 & 27.0 & 26.0 &    48.0 & 48.0    & 68.0 & 68.0 \\
   p-10 & 0.2 & 24.0 & 23.0 &    41.0 & 39.0    & 44.0 & 43.0 \\
   p-10 & 0.5 & 20.0 & 18.0 &    35.0 & 33.0    & 39.0 & 36.0 \\
  \midrule
   p-20 & 0.1 & 51.0 & 49.0 &    95.0 & 93.0    & 96.0 & 96.0 \\
   p-20 & 0.2 & 53.0 & 49.0 &    86.0 & 82.0    & 80.0 & 80.0 \\
   p-20 & 0.5 & 41.0 & 33.0 &    66.0 & 61.0    & 50.0 & 46.0 \\
  \midrule
   p-40 & 0.1 & 110.0 & 103.0 &    197.0 & -    & 176.0 & 174.0 \\
   p-40 & 0.2 & 96.0 & 92.0 &    167.0 & -    & 162.0 & 162.0 \\
   p-40 & 0.5 & 71.0 & 69.0 &    127.0 & 123.0    & 92.0 & 88.0 \\
\bottomrule
\end{tabular}
\end{center}
\end{table*}

\subsection{\rooms Domain}

The \rooms domain is the one defined in Table~\ref{tab:actions}. Opportunities are objects being held by the robot
because another agent gave them or objects prepared before they are needed by other robots. In order to simulate a real
environment, every time the robot executes some action, the perception system discovers five new objects in the room it
is operating at. This is only relevant for \replan, since \clo focuses only on the found opportunities. The initial
length of the plans were: 14, 29, 59 and 119 actions, respectively.

\subsection{\dialog Domain}

The \dialog domain consists of a robot that operates in an open environment and has to approach humans to ask them a
sequence of questions. It uses two actions ({\tt move} and {\tt ask}) and four predicates ({\tt at-robot, at-human,
  asked} and another one for defining the dependency between questions, {\tt depends}). In the initial state, the robot
did not ask any question to any available human, and in the goals, it should have asked five questions to all these
humans. Opportunities arise because, sometimes, humans provide information on future questions in the surveys when
answering previous questions.  In this domain, the robot's perception system sees five new humans in their surrounding
every time it executes an action in the plan. The number of actions of the initial plans were: 30, 60, 120 and 240
actions, respectively.

\subsection{\cooking Domain}

The \cooking domain consists of a robot that has to cut some specific quantities of ingredients in a restaurant. It uses
two actions ({\tt get} and {\tt cut} ingredient) and two predicates ({\tt has-ingredient}, and {\tt has-cut}, to define
how many units of each ingredient the robot has cut). In the initial state, the robot did not cut any ingredient, and in
the goals, it should have have cut a given amount of units of those ingredients.  Opportunities might arise given that
the robot might expect that the ingredients can yield a given quantity of cut units, but there are ingredients of
varying size that will yield different quantities when cut. Also, there might be other agents cutting ingredients
unexpectedly generating the required cut units. In this domain, the robot perceives five more ingredients every time it
executes an action. The number of actions in the initial plans were: 35, 85, 185 and 385 actions, respectively.

\subsection{Results}

Figure~\ref{fig:exp} shows the total time employed by \clo and \replan (in seconds) in the three domains: (a) \rooms,
(b) \dialog and (c) \cooking.  The problems are on the x-axis, while the different lines correspond to the probabilities
of opportunity appearance. The graph for time for \replan in the \dialog domain stops at p-20, since it cannot solve
the p-40 problem due to the time constraint.  Table~\ref{tab:num-actions} shows the number of executed actions in the
three domains. A dash means the task could not be solved in the allotted time.

\subsection{Discussion}

We can draw the following conclusions that are common for all domains. The computational effort measured in terms of
computation time or number of nodes expanded of \replan can be up to two orders of magnitude bigger than that of
\clo. Changes in the environment in most robotic domains occur at each time step. Since \replan has to replan every time
there is a change in the environment, the total planning time is a burden for this approach compared to the short time
required to perform repairs in the plan. This is true even in domains, such as the minimalist versions used in the
experiments, where the planning time is really small (less than one second per planning step in most replanning
episodes). This effect is most noticeable in the \dialog domain where \replan cannot solve the hardest task in the
allotted time in two out of the three configurations.

The computational effort is also affected by the probability of appearance of an opportunity, especially in the case of
\replan. The higher the probability, the higher the reduction in computational effort, since the length of execution
(and thus planning) episodes decreases. Also, since \clo provides the monitoring system with a reduced set of
propositions to look at, the perception system only has to focus on those. Therefore, the perception cost gets reduced
accordingly.

In terms of quality of solutions, the results are similar for both configurations. The number of executed actions of
\replan is slightly less than \clo, given that replanning can potentially find improvements on the new plan that cannot
be found by our repair approach. And, again, there is a huge influence on how many opportunities appear. When many
opportunities appear (high probability of appearance), the number of executed actions can decrease to half the number of
executed actions with less opportunities.

\section{Conclusions and Future Work}

In this paper, we present three main contributions: definition and computation of opportunities on dynamic predicates;
detection of opportunities during execution; and repairing a plan when some opportunity arises. Experimental results
show the benefits that using this approach brings to three domains that cover a wide range of representative robotic
tasks.  As future work, we would like to study other alternatives to improve the repair mechanism.  Besides, this work
checks the opportunities at every execution step. However, some opportunities may be only, or more relevant, at some
execution step. Furthermore, we do not consider perception cost of each type of opportunity.  In our future work, we
will continue to enrich the model of interleaving planning and execution, bringing to execution knowledge about the
potential value of opportunities, but also weighing the cost, frequency, and timing of opportunity perception.

\section*{Acknowledgments}

This paper was prepared for information purposes by the Artificial
Intelligence Research group of JPMorgan Chase \& Co. and its
affiliates (``JP Morgan''), and is not a product of the Research
Department of JP Morgan.  JP Morgan makes no representation and
warranty whatsoever and disclaims all liability, for the completeness,
accuracy or reliability of the information contained herein.  This
document is not intended as investment research or investment advice,
or a recommendation, offer or solicitation for the purchase or sale of
any security, financial instrument, financial product or service, or
to be used in any way for evaluating the merits of participating in
any transaction, and shall not constitute a solicitation under any
jurisdiction or to any person, if such solicitation under such
jurisdiction or to such person would be unlawful.  This work has been
partially funded by FEDER/Ministerio de Ciencia, Innovaci\'on y
Universidades - Agencia Estatal de Investigaci\'on,
TIN2017-88476-C2-2-R.

\bibliographystyle{IEEEtran.bst}
\bibliography{general-daniel,daniel,manuela}

\end{document}